\documentclass[runningheads]{llncs}

\usepackage{eccv}

\usepackage{eccvabbrv}

\usepackage{amsmath}
\usepackage{amssymb}
\usepackage{cancel}
\usepackage{subcaption}
\usepackage{multirow}
\usepackage{xcolor}
\usepackage{mathtools}
\usepackage{comment}

\definecolor{es-blue}{rgb}{0,0.4,0.8}

\usepackage{pifont}
\newcommand{\cmark}{\ding{51}}%
\usepackage{graphicx}
\usepackage{booktabs}

\usepackage[accsupp]{axessibility}  % Improves PDF readability for those with disabilities.

\usepackage{hyperref}

\usepackage{orcidlink}

\begin{document}

\titlerunning{Feature Splatting}

% ---------------------------------------------------------------
\title{Feature Splatting: Language-Driven Physics-Based Scene Synthesis and Editing}

% TODO FINAL: Replace with your author list. 
% Include the authors' OCRID for the camera-ready version, if at all possible.
\author{Ri-Zhao Qiu\inst{1} \and
Ge Yang\inst{2,3} \and
Weijia Zeng\inst{1} \and
Xiaolong Wang\inst{1}}

% TODO FINAL: Replace with an abbreviated list of authors.
\authorrunning{Qiu et al.}
% First names are abbreviated in the running head.
% If there are more than two authors, 'et al.' is used.

% TODO FINAL: Replace with your institution list.
\institute{University of California San Diego
\and
Massachusetts Institute of Technology
\and
Institute for Artificial Intelligence and Fundamental Interactions \\
\color{es-blue}{\texttt{\url{https://feature-splatting.github.io}}}
}

\maketitle

\begin{abstract}
  Scene representations using 3D Gaussian primitives have produced excellent results in modeling the appearance of static and dynamic 3D scenes. Many graphics applications, however, demand the ability to manipulate both the appearance and the physical properties of objects. We introduce \textit{Feature Splatting}, an approach that unifies physics-based dynamic scene synthesis with rich semantics from vision language foundation models that are grounded by natural language. Our first contribution is a way to distill high-quality, object-centric vision-language features into 3D Gaussians, that enables semi-automatic scene decomposition using text queries. Our second contribution is a way to synthesize physics-based dynamics from an otherwise static scene using a particle-based simulator, in which material properties are assigned automatically via text queries. We ablate key techniques used in this pipeline, to illustrate the challenge and opportunities in using feature-carrying 3D Gaussians as a unified format for appearance, geometry, material properties and semantics grounded on natural language.
\end{abstract}

\section{Introduction}
\label{sec:intro}

What does a falling leaf know about autumn? A video of this moment, had someone captured it, may include a delicate dance with the autumn breeze. Although invisible to the camera, the shape of the wind is carved out by the leaf's swirling path, and thus, became fully visible to our eyes. This technique of using motion to make the invisible visible lies behind many artistic manipulations of images and movies. A leaf that falls straight down versus a leaf that floats across a busy street bouncing up and down tells very different stories. In the latter case we would intuitively reason about the presence of the wind despite not directly feeling its touch on our skins; and we would be drawn in --- as if it were us who are being carried away.

\begin{figure*}[t]
    \centering
    \includegraphics[width=\linewidth]{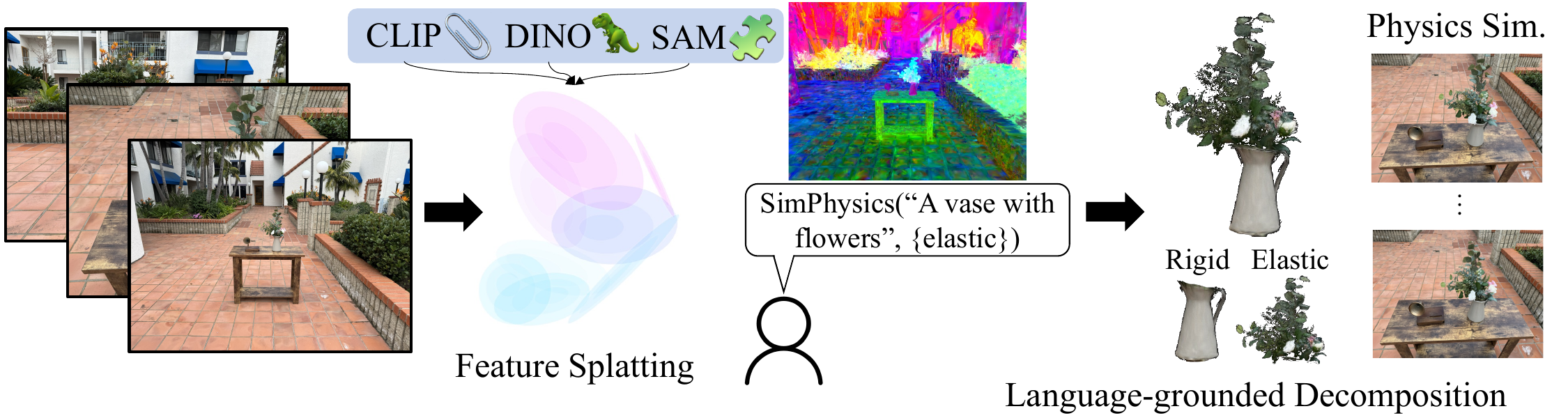}
    \caption{\textbf{Feature Splatting.} An overview of the language-grounded scene physics editing pipeline. Given input images, feature splatting optimizes for a unified Gaussian representation that contains the geometry, texture, and semantics of the scene using features from large-scale 2D vision models~\cite{radford2021-CLIP,oquab2023-dinov2,kirillov2023-segmentanything}. With open-vocabulary scene decomposition, feature splatting segments an object and automatically determines the physical properties of components within the object. In this example, a user gives a query `a vase with flowers'. Feature splatting extracts the vase with flowers in the scene, and further decomposes it into rigid and non-rigid parts, creating a dynamic scene of flowers swaying in the wind. (Best viewed in videos on project website).}
    \label{fig:teaser}
    \vspace{-1.5em}
\end{figure*}

In this work, we present Feature Splatting, a way to semi-automatically synthesize dynamic scenes from an otherwise static 3D capture, where we use open-text language queries to jointly manipulate the appearances and assign material properties governing the dynamic interactions. We do so by augmenting existing point-splatting methods that use 3D Gaussians as geometric primitives~\cite{yifan2019-pointsplatting,Zwicker2001-surface-splatting,kerbl2023-gaussiansplatting} with additional view-invariant features sourced from vision, and vision-language foundation models~\cite{kirillov2023-segmentanything,radford2021-CLIP,oquab2023-dinov2} within the same analysis-by-synthesis pipeline. We further extend to physics-based dynamic scene synthesis, by augmenting the static capture with particle-based interaction, where material categories and properties are assigned semi-automatically via natural language queries. The resulting format unifies photo-realism, rich semantics, and physics-based dynamic synthesis in a single format.

Both the extension to feature-carrying 3D Gaussians, and the physics-based dynamic synthesis involve unexpected technical challenges. To begin with, we found that Gaussian primitives from~\cite{kerbl2023-gaussiansplatting} effectively shares the same interpolation kernel for both the geometry and the radiance, whereas the 2D feature maps we source from reference camera views are both low-resolution and noisy. A naive distillation pipeline similar to NeRF-based methods~\cite{kobayashi2022-DFF} lead to poor results with a lot of high-frequency feature noise (see Fig.~\ref{fig:feature_comparison}). We address this problem by introducing a novel method for extracting the feature maps, and a procedure for distilling them. On the dynamic synthesis side, we propose ways to in-fill existing static 3D Gaussian captures for volume-dependent physical effects, and ways to transform the Gaussian primitives under significant deformation that is distinct from, and perform better than prior approaches.

The scene modeling and synthesis pipeline, Feature Splatting, contains rich semantic priors that enables easy, language-driven edits involving decomposing a static 3D capture and associating each constituent with material properties for physics-based dynamic synthesis. Feature-carrying 3D Gaussians serve as a unified representation for appearance, semantics, geometry, and physics. 

Our contribution is threefold:
\begin{enumerate}
    \item {\bf A method}, feature splatting, to augment static scenes with semantics and language-grounded physically realistic movements.
    \item {\bf Techniques to the algorithmic and systems challenges} towards a unified representation: {\bf an MPM-based~\cite{hu2019-taichi} physics engine} that is adapted to Gaussian-based representation; {\bf a novel way to fuse features} from multiple foundation vision 2D models for accurate decomposition.
    \item {\bf A demonstration} that feature splatting is an excellent editing tool that enables {\it automatic} language-grounded scene editing.
\end{enumerate}

\section{Related Work}
\label{sec:related}

\paragraph{Novel View Synthesis.} Recently, the computer graphics community has seen a growing interest in using differentiable rendering to optimize scene representations for novel view synthesis~\cite{mildenhall2021-nerf,wang2023-f2nerf,barron2021-mipnerf,fridovich2022-plenoxels,muller2022-instantNGP,kerbl2023-gaussiansplatting}. Recent techniques can be categorized into implicit methods~\cite{mildenhall2021-nerf,wang2023-f2nerf,barron2021-mipnerf,fridovich2022-plenoxels,muller2022-instantNGP} and explicit methods~\cite{kerbl2023-gaussiansplatting}. The representative work of implicit methods is Neural Radiance Fields (NeRFs), which was proposed by~\cite{mildenhall2021-nerf} and learns a neural network to predict the radiance of the scene. Later work in NeRFs advance the technique by accelerating the training process~\cite{muller2022-instantNGP}, alleviating the artifacts at the boundary of the scene~\cite{wang2023-f2nerf}, and addressing texture aliasing~\cite{barron2021-mipnerf}. For explicit scene representation methods, Gaussian Splatting (GS)~\cite{kerbl2023-gaussiansplatting} is a recent method that achieves both fast training time and high rendering quality. GS represents the scene as a collection of Gaussians that can be explicitly manipulated. Our work is built on GS without compromising its novel view synthesis capability, where we leverage the explicit representations in GS to connect it with particle-based mechanics and feature distillation.

\paragraph{Scene Editing with Distilled Feature Fields.} Various work has proposed ways to make edits with NeRFs. Existing works mainly focus on manipulating the appearance~\cite{kobayashi2022-DFF,jambon2023-nerfshop,haque2023-instructnerf,li2023-climatenerf}. For instance, Distiled Feature Fields (DFF~\cite{kobayashi2022-DFF}) performs appearance editing via zero-shot open-text segmentation, in which it uses knowledge distillation to embed features from 2D foundation vision models. During rendering, DFF decomposes the scene by relating language query and distilled features to segment affected volumes. Appearance editing such as color change or removal can then be performed on segmented objects. NeRFShop~\cite{jambon2023-nerfshop} proposes an interactive pipeline that allows user input to make geometric modifications to NeRF. Instruct-NeRF2NeRF~\cite{haque2023-instructnerf} takes a different approach to scene editing. Instead of injecting the rendering process, Haque et al.~\cite{haque2023-instructnerf} proposes to use an off-the-shelf 2D editing method~\cite{brooks2023-instructpix2pix} to modify the images used to train NeRFs. Most closely related to our work, ClimateNeRF~\cite{li2023-climatenerf} proposes to inject the rendering process with physical simulation to simulate different weather effects. However, due to the intrinsic limitation of implicit scene representation, ClimateNeRF only supports modifying the ray marching progress in neural rendering, which limits it to simulating ray reflection, refraction, and diffraction for weather effects. In comparison, our method uses explicit representations to support object-centric physical simulation, which has much broader potential applications.

\paragraph{Concurrent Work.} During the preparation of this manuscript, a few concurrent works appeared, that also studied ways to edit or simulate dynamic scenes via 3D Gaussians. The majority of these works focus solely on segmenting the scene via language-grounded semantics~\cite{shi2023-legaussian,ye2023-gaussian-grouping,zhou2023-feature3dgs,kirillov2023-segmentanything}. Most closely related to our work, Xie \etal~\cite{xie2023-physgaussian}, propose a similar rendering-simulation pipeline that also uses material point methods (MPM). Semantic grounding in natural language is not part of their proposal, and they manually select and assign material properties to the Gaussians. The way these two works handle the rotation of Gaussians is also different. We include results that show rotation from deformation gradients, proposed in~\cite{xie2023-physgaussian}, which fails to maintain rendering quality when deformation is large. Along the line of works that perform segmentation, Feature3DGS~\cite{zhou2023-feature3dgs} is most relevant to our work for its feature distillation designs to fuse 2D reference features using priors from multiple foundation models. We consider feature distillation one component of a larger simulation, rendering, and editing pipeline, with systems optimization techniques that speed up training by 30\%.

\begin{figure*}[t]
    \centering
    \includegraphics[width=\linewidth]{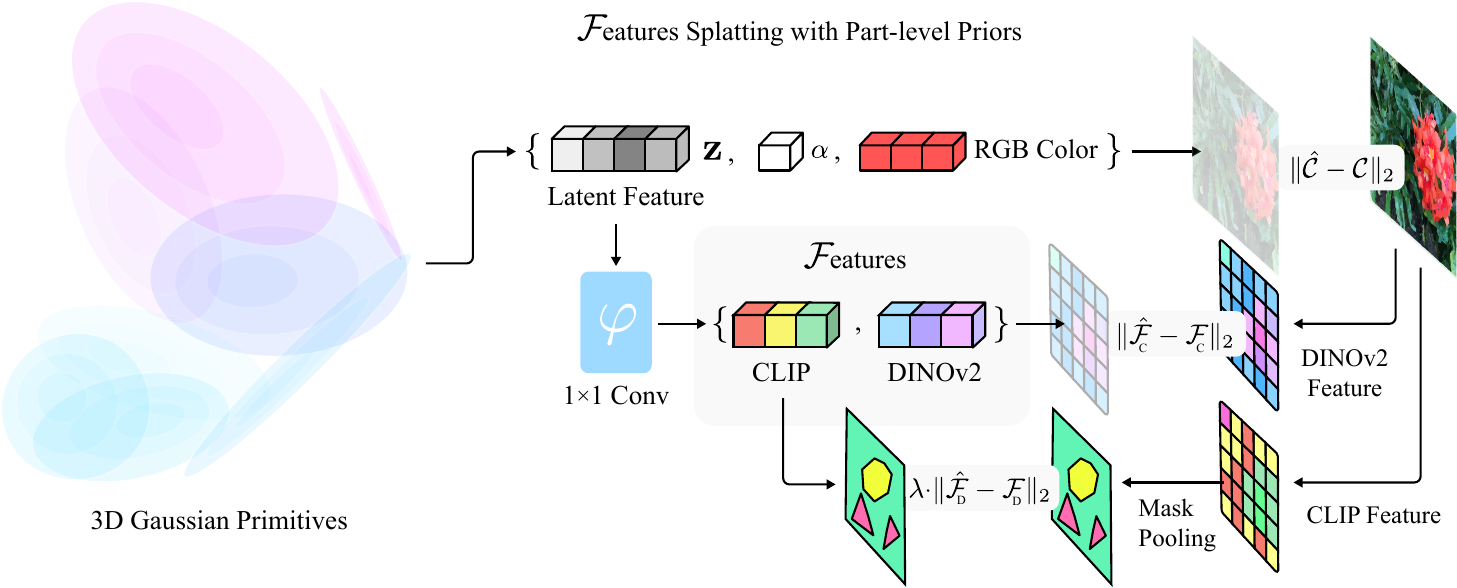}
    \caption{\textbf{Feature Splatting.} Raw CLIP features are noisy and low-resolution. We improve the quality of the feature maps by pooling within part-level masks produced by the Segment Anything Model (SAM~\cite{kirillov2023-segmentanything}). Jointly modeling features from DINOv2~\cite{oquab2023-dinov2} and CLIP is an optional regularization that offers minor additional improvements.
    }
    \label{fig:method_overview}
    \vspace{-1em}
\end{figure*}

\section{Language-Driven Physics-Based Synthesis and Editing}

We present three key components of our dynamic scene synthesis and editing pipeline: First, a way to distill rich semantic features from vision-language models into 3D Gaussians. Second, a way to decompose a scene into key constituents using open-text queries. Finally, a way to ground the material properties via language, as part of a physics-based dynamic scene synthesis procedure.

\subsection{Differentiable Feature Splatting}
\label{sec:feature_rendering}

Point and surface splatting methods~\cite{yifan2019-pointsplatting,Zwicker2001-surface-splatting} represent a scene explicitly via a mixture of 2D or 3D Gaussian primitives. In the case of Gaussian Splatting~\cite{kerbl2023-gaussiansplatting}, the geometry is represented as a collection of 3D Gaussian, each being the tuple \(\{\mathcal X, \mathbf\Sigma\}\) where $\mathcal{X} \in \mathbb{R}^3$ is the centroid of the Gaussian and $\Sigma$ is its covariance matrix in the world frame. This gives rise to the probability density function%
\begin{equation}
    G(\mathcal{X}, \Sigma) = \exp{- \frac 1 2 \mathcal{X}^\top \Sigma^{-1} \mathcal{X}}\,.
\end{equation}%
To ensure \(\Sigma\)'s positive semi-definiteness during optimization, it is common practice to decompose it into a scaling matrix $\mathbf{S}$ and a rotation matrix $\mathbf{R}$ via \(\Sigma = \mathbf{R} \mathbf{S} \mathbf{S}^\top \mathbf{R}^\top\).  The color information in the texture is encoded with a spherical harmonics map \(\mathbf c_i = \mathrm{SH}_{\phi}(\mathbf d_i)\), which is conditioned on the viewing direction $\phi$. 

\paragraph{Feature Splatting.} Feature Splatting appends an additional vector $\mathbf{f}_i \in \mathbb{R}^d$ to each Gaussian, which is rendered in a view-independent manner because the semantics of an object shall remain the same regardless of view directions. The rasterization procedure starts with culling~\cite{kerbl2023-gaussiansplatting} the mixture by removing points that lay outside the camera frustum. The remaining Gaussians are projected to the image plane according to the projection matrix \(\mathbf{W}\) of the camera. This projection also induces the following transformation on the covariance matrix $\Sigma$:%
\begin{equation}
    \Sigma^{'} = \mathbf{J}\,\mathbf{W}\,\Sigma\,\mathbf{W^\top J^\top}\,,
\end{equation}
where $\mathbf{J}$ is the Jacobian of the projection matrix \(\mathbf{W}\). We can then render both the color and the visual features with the splatting algorithm:%
\begin{equation}
    \{\hat{\mathbf{F}}, \hat{\mathbf{C}\}} = \sum_{i \in N} \{\mathbf{f}_i, \mathbf{c}_i\}\cdot \alpha_i\, \prod_{j = 1}^{i - 1} (1 - \alpha_j)\,,
\end{equation}%
where $\alpha_i$ is the opacity of the Gaussian conditioned on $\Sigma^{'}$ and the indices $i\in N$ are in the ascending order determined by their distance to the camera origin.

\begin{figure*}[t!]%
    \centering
    \includegraphics[width=\linewidth]{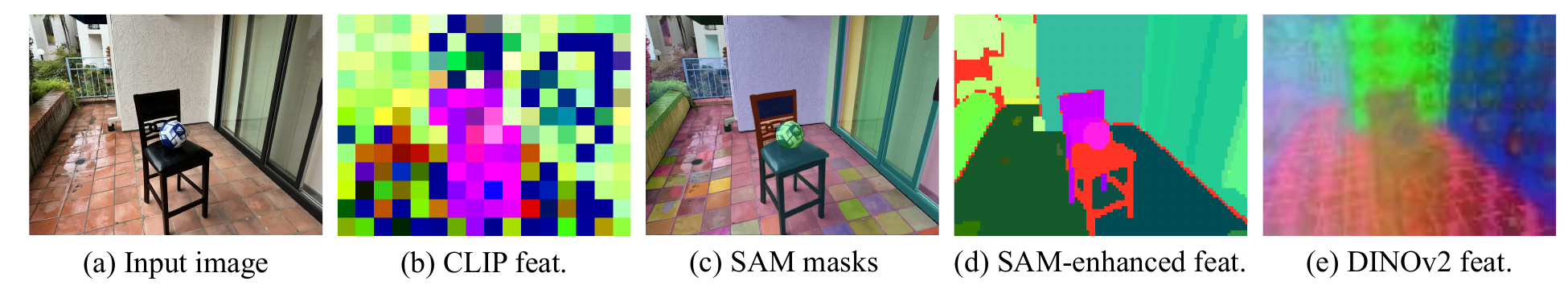}%
    \vspace{-1em}
    \caption{\textbf{Raw and Enhanced Feature Maps.} CLIP features contain view-dependent noise that degrades the feature splats~\cite{radford2021-CLIP}. We mask-pool with masks produced by SAM~\cite{kirillov2023-segmentanything}, and regularization through joint modeling of DINOv2 features~\cite{oquab2023-dinov2} to improve its quality. Color corresponds to top three PCA vectors.}
    \label{fig:feature_comparison}
\end{figure*}

\paragraph{Systems Considerations.} Na\"ively extending the color rasterizer in vanilla GS~\cite{kerbl2023-gaussiansplatting} to rasterize high-dimensional features causes expensive training time. We performed an in-depth analysis and found that the main bottleneck lies in the memory access pattern, which we addressed by designing custom CUDA kernels. The details are presented in the appendix.

\paragraph{Improving Reference Feature Quality Using Part-Priors.}

Our differentiable feature splatting is a generic method, where the resulting features depend on the reference features. Though CLIP~\cite{radford2021-CLIP} is a commonly used 2D vision models to obtain language-aligned features, na\"ively splatting CLIP features results in low-quality 3D features given the coarseness of the CLIP features (see Fig.~\ref{fig:feature_comparison} and Fig.~\ref{fig:rendering_regularization}). This issue is less pronounced in NeRF-based methods~\cite{kobayashi2022-DFF,kerr2023-lerf} because the continuous representations of NeRF serves as an implicit regularization. On the other hand, explicit representations such as GS~\cite{kerbl2023-gaussiansplatting} have no such regularization and are prone to overfit to noises inherited from coarse reference feature maps.

We propose a way to improve the quality of the Gaussian features using object priors from DINOv2~\cite{oquab2023-dinov2} and the Segment Anything Model (SAM)~\cite{kirillov2023-segmentanything}. Consider an input image. We first use SAM to generate a set of part-level masks \(\{\mathbf{M}\}\) (see Fig.~\ref{fig:feature_comparison}\textcolor{red}{c}). For a given binary mask $\mathbf{M}$ and the coarse CLIP feature map $\mathbf{F}_{C}$, we use Masked Average Pooling (MAP) to aggregate a single feature vector%
\begin{equation}
    w = \textsc{MAP}(\mathbf{M}, \mathbf{F}_{C}) = \frac{\sum_{i \in \mathbf{F}_{C}} \mathbf{M}(i) \cdot \frac{\mathbf{F}_{C}(i)}{\left| \left| \mathbf{F}_{C}(i) \right| \right|}}{\sum_{i \in \mathbf{F}_{C}} \mathbf{M}(i)}\,,
\end{equation}%
where $i$ is a pixel coordinate in the feature map. $w$ is then assigned to all pixels that are within the part segmentation. If a pixel belongs to multiple parts, the pixel feature is obtained by averaging all relevant part features. This gives us a SAM-enhanced CLIP feature map (shown in Fig.~\ref{fig:feature_comparison}\textcolor{red}{d}).

To further reduce the possibility of overfitting, we introduce a shallow MLP with two output branches that takes in the rendered features $\hat{\mathbf{F}}$ as intermediate features. The first branch renders the DINO~\cite{Caron2021-dino} feature $\hat{\mathbf{F}}_D$ for its coherent part-level semantics (see Fig.~\ref{fig:feature_comparison}\textcolor{red}{e}), and the second branch renders the CLIP~\cite{radford2021-CLIP} features $\hat{\mathbf{F}}_C$.%
\begin{align}
    \hat{\mathbf{F}}_{C}, \hat{\mathbf{F}}_{D} &= \textsc{MLP}(\hat{\mathbf{F}})\,,
\label{eq:dino_regularization}
\end{align}%
where $\hat{\mathbf{F}}_C$ is supervised using the SAM-enhanced CLIP feature map with cosine loss, and $\hat{\mathbf{F}}_D$ is supervised using the DINOv2~\cite{oquab2023-dinov2} feature map using cosine loss. We scale the CLIP term in the joint loss \( \mathcal{L}_\text{CLIP} + \lambda \cdot \mathcal{L}_\text{DINO}\,\) with $\lambda = 0.1$, so the optimization focuses on language grounding and treat DINO features as a mild smoothing term. In sum, this would give us Gaussians with regularized CLIP features.

\subsection{Language-guided Scene Decomposition}
\label{sec:editing_main_sec}

We first perform object- and part-level open-vocabulary scene decomposition, where we take language queries to coarsely select objects for editing ({\it e.g., a vase with flowers}) and feature splatting automatically decouples object components for simulation ({\it e.g., vase is rigid and stems of flowers are elastic}).
We do so by identifying Gaussians whose CLIP features more closely align with positive queries over negative queries.

More specifically, given a positive vocabulary ({\it e.g., \(L^+=\) `bulldozer'}) and generic negative vocabularies \(\{L^-_i\}_{i=0}^N\) ({\it i.e., `objects' and `things'}), we use frozen CLIP text encoder to obtain text embeddings for every vocabulary. Then we follow standard CLIP practice~\cite{radford2021-CLIP} and compute pair-wise cosine similarity between rasterized CLIP feature of every Gaussian and the text embeddings. A temperatured softmax is then applied to similarities to obtain probability distribution, where we select Gaussians whose similarity to $L^+$ passes a certain threshold $\tau = 0.6$ as the foreground object. We include segmentation results using negative text-queries in the appendix.

\subsubsection{Basic Editing Primitives.} Feature splatting supports various basic editing primitives, which can be easily composed to achieve more complex behavior. Let $\{\hat{\mathcal{X}}, \hat{\mathbf\Sigma}\} \subseteq \{\mathcal X, \mathbf\Sigma\}$ be the set of Gaussians selected to be edited. We briefly describe how basic editing primitives are implemented.

\begin{itemize}
    \item {\bf Object Removal.} Objects are removed by simply removing selected Gaussians $\{\mathcal X, \mathbf\Sigma\} \coloneqq \{\mathcal X, \mathbf\Sigma\} \setminus \{\hat{\mathcal{X}}, \hat{\mathbf\Sigma}\}$.
    \item {\bf Translation.} Given a displacement vector $\mathbf{b}_1 \in \mathbb{R}^3$, objects can be easily displaced by shifting the Gaussian centroid $\hat{\mathcal{X}} \coloneqq \hat{\mathcal{X}} + \mathbf{b}_1$.
    \item {\bf Rotation.} Given a rotation matrix $\mathbf{R}_1 \in \mathbf{SO}(3)$, we modify the covariance of selected Gaussians as $\hat{\Sigma} \coloneqq \mathbf{R}_1\hat{\mathbf{R}} \hat{\mathbf{S}} \hat{\mathbf{S}}^\top \hat{\mathbf{R}}^\top \mathbf{R}_1^\top$.
    \item {\bf Scaling.} Given an axis-aligned scaling vector $\mathbf{s}_1 \in \mathbb{R}^3$, we can scale the size of objects via $\hat{\mathcal{X}} = \mathbf{s}_1\hat{\mathcal{X}}, \quad \hat{\Sigma} \coloneqq \hat{\mathbf{R}} (\mathbf{s}_1\hat{\mathbf{S}}) (\mathbf{s}_1\hat{\mathbf{S}})^\top \hat{\mathbf{R}}^\top$.
    % \item {\bf Transparency.} Given an opacity scaling coefficient $\beta \in [0, 1]$, we scale the opacity coefficient $\hat{\mathbf{\alpha}} \coloneqq \beta \hat{\mathbf{\alpha}}$.
\end{itemize}

\subsection{Language-Driven Physics Synthesis.}

Feature splatting can automatically choose physical properties for simulation, estimate collision surfaces, and predict gravity for simulation. Based on the explicit representation, we extend the material-point method (MPM)~\cite{sulsky1994-MPM} using Taichi~\cite{hu2019-taichi} to augment objects with various physical properties.

\paragraph{Decoupling Objects for Simulation.} We construct a set of vocabularies for common rigid materials such as ({\it e.g., wood, ceramic, and steel}), which can be expanded with optional user inputs. Given a selected multi-part object ({\it e.g., a vase with flowers}), we perform another round of CLIP similarity comparison to select particles within the object that are aligned more closely to these materials. The selected particles are considered to be rigid during simulation.

\paragraph{Language-grounded Collision Surface Estimation.} We feed a canonical set of queries including common planar objects ({\it e.g., floor and tabletop}) into the scene decomposition pipeline described above to obtain Gaussians of these objects. Then we apply RANSAC~\cite{fischler1981-RANSAC} to estimate these plane geometry, which are used as collision surfaces in physics simulation. The gravitational directional vector is estimated as the normal vector of plane geometry of ``floor''.

\paragraph{Taichi MPM for gaussians.} Given an selected object with rigid parts, collision surfaces, and gravity, we may simulate physics with particle mechanics. While MPM methods~\cite{sulsky1994-MPM,hu2019-taichi} can be directly applied to Gaussians by treating the Gaussian centroids $\mathcal{X}$ as point clouds, doing so results in unsatisfactory quality. Parallel to findings by Xie \etal~\cite{xie2023-physgaussian}, we find that simulating and displacing only the surface centroids leads to issues such as 1) object collapsing on contact with collision surfaces due to lack of internal support and 2) undesired artifacts when objects undergo deformation.

We build our gaussian-oriented material-point method (GS-Taichi-MPM) based on Taichi~\cite{hu2019-taichi}, which supports realistic physical simulation of various types of materials ({\it e.g., rigid, elastic, granular (sand), and liquid}). Our method goes beyond simple point-based physics simulation and makes use of gaussian-specific information, such as isotropic opacity and Gaussian covariances for volume preservation and covariance modification during deformation, to address the two aforementioned challenges.

\begin{itemize}

\item \textbf{Implicit Volume Preservation.} One of the open challenges in simulating physics from a few real-world images is volume preservation. For instance, without volume preservation, a volleyball simulated to hit the ground would collapse upon collision. Hence, we propose an implicit volume preservation technique using the opacity and covariances of gaussians. Specifically, we first densify surface points by sampling points on disks of surface gaussians using covariance and opacity information following~\cite{tang2023-dreamgaussian}. With densified surface points, we then fill transparent supporting particles extending from the centroid of the object to the surface. The transparency of filled particles is intended to enforce identical rendering quality at $T=0$ for a smooth and continuous transition from static scene to dynamics.

\item \textbf{Estimating Rotation.} When the object undergoes deformation, artifacts may manifest without correcting the rotational component of the covariance matrix. Notably, Xie \etal~\cite{xie2023-physgaussian} noted a similar issue and attempted to use the deformation gradient $F$ of MPM to update Gaussian covariances. With slight abuse of the $\Sigma$ notation, let $\mathbf{F} = U \Sigma V^\top$ be the per-particle defomration gradient and its SVD factorization, rotation from deformation can be estimated as $\mathbf{R}_1 = VU^\top$. However, deformation gradients capture mostly local deformation, and we empirically find that this approach fails when elastic objects undergo large deformation (see Fig.~\ref{fig:ours_rotation}).

We propose to estimate the rotation matrix for {\it elastic} objects using normals. Since Gaussian splatting does not estimate normals due to challenges in sparse reconstruction~\cite{kerbl2023-gaussiansplatting}, we use an alternative NN-based approach. In particular, for every Gaussian to be simulated, we find two of its nearest neighbors. The centroids of these three Gaussians form a plane, and we use the rotation of the plane normal throughout the object dynamics as a proxy. Compared to rotation from deformation, our approach yields fewer artifacts when objects undergo large deformation (see Fig.~\ref{fig:ours_rotation}).
\end{itemize}

\begin{figure*}[t]%
    \centering
    \includegraphics[width=0.96\linewidth]{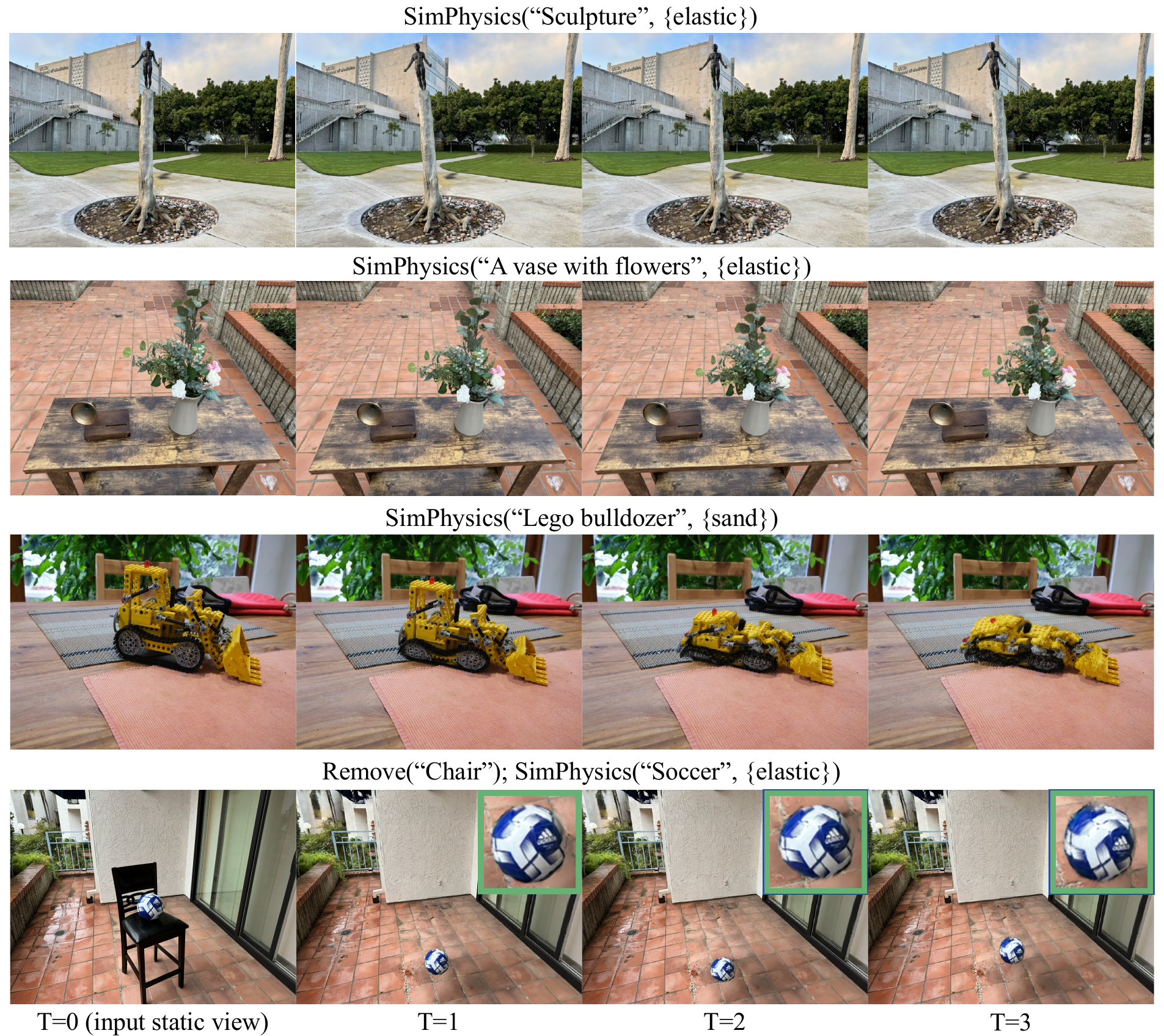}%
    \caption{
    \textbf{Physics-based Dynamic Scene Synthesis.}
    Rich semantic features in Feature Splatting enable semi-automatic assignment of material properties for synthesizing dynamic scenes from a single static 3D capture. We can use simple text queries to manipulate the physical property of specific objects and materials. From top to bottom: changing the elasticity, turning solid into granular material, modeling volume-dependent deformation in a falling volleyball.
    \bf{For the best illustration with animations and moving cameras, please refer to videos on the project website.}
    }
    \label{fig:physics_simulation_main}
    \vspace{-2em}
\end{figure*}

\section{Experiments}

We provide qualitative results on language-driven editing and dynamic synthesis, and quantitative comparisons against radiance-field based approaches when applicable.

\textbf{Datasets.} We use the deep blending dataset~\cite{hedman2018-dbdataset} and the Mip-NeRF360~\cite{barron2022-mipnerf360} dataset, where we compute the camera intrinsics, extrinsic, and sparse point clouds using colmap~\cite{schonberger2016-colmap-sfm,schoenberger2016-colmap-mvs}. Following LERF~\cite{kerr2023-lerf}, we evaluate the localization capability of feature splatting with 72 objects using language query in five localization scenes from LERF~\cite{kerr2023-lerf}. Finally, we collect a custom dataset to demonstrate the capability of our method to perform physical simulations. The custom data sequences are captured using the main camera of an iPhone 15 Pro with intrinsics, extrinsic, and initial sparse point cloud computed by colmap. We plan to release the custom dataset to aid future research.

\textbf{Metrics.} For physics editing, since there are no ground truth images or previous baselines to compare to, we focus on qualitative comparisons. To evaluate the localization accuracy, we follow the same evaluation protocol as LERF~\cite{kerr2023-lerf} and compute localization accuracy on testing views. We also evaluate how much different components in our systems contribute to training efficiency. Finally, we report PSNR and cosine feature rendering loss on training and validation views to validate necessity for fusing multiple models and show that feature splatting does not interfere with rendering quality.

\begin{figure*}[t]
    \centering
    \includegraphics[width=\linewidth]{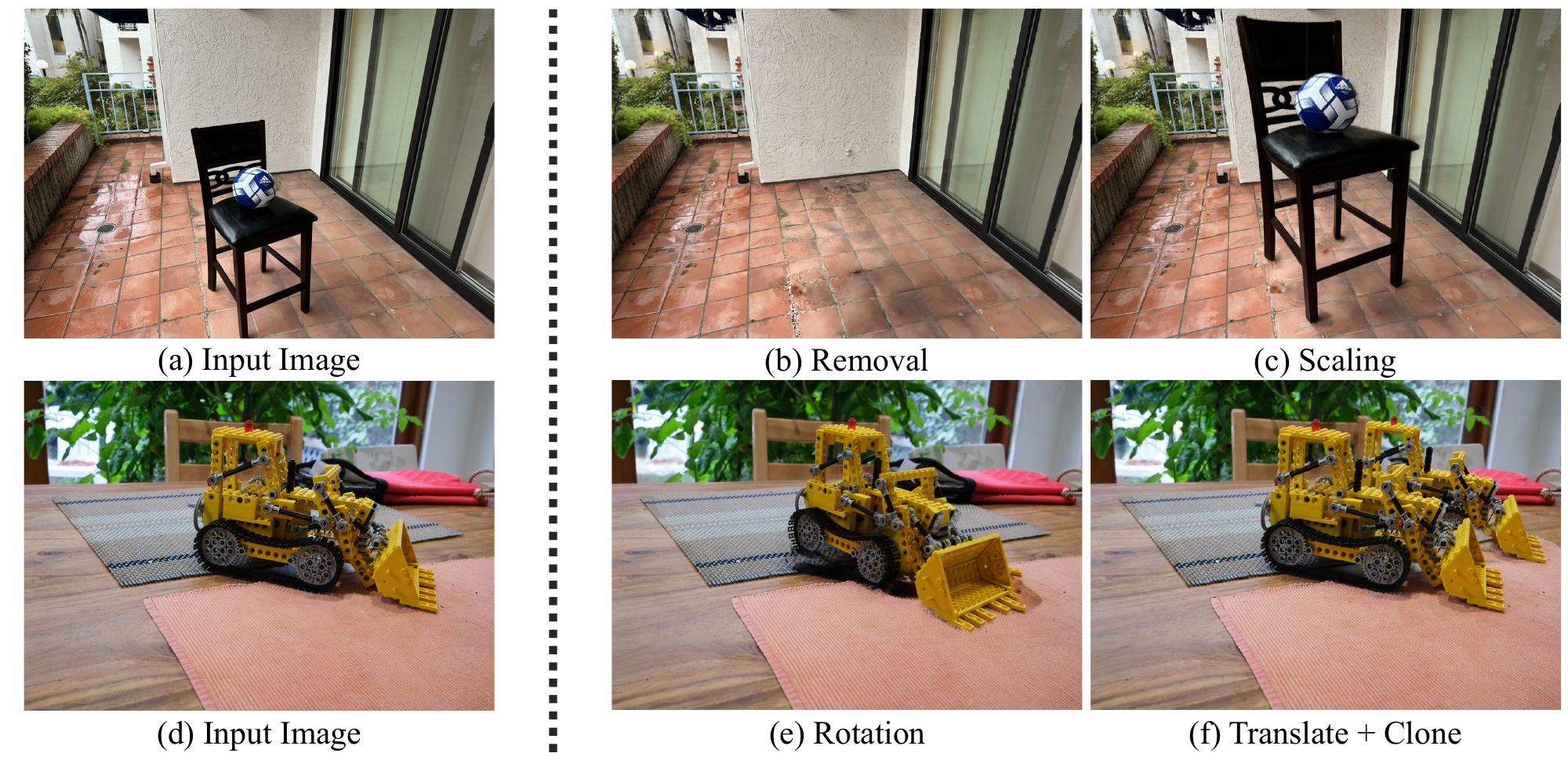}
    \caption{\textbf{Feature Splatting Editing Primitives.}: We can remove, scale, rotate, translate, and clone objects in the scene using language.}
    \label{fig:editing_primitives_viz}
    \vspace{-1em}
\end{figure*}

\subsection{Dynamic Scene Synthesis Results}

We present synthesized dynamic scenes that involves elastic deformation, loose, granular materials, and volume-preserving deformations with a volleyball in Figure~\ref{fig:physics_simulation_main}. These results are produced by selecting from a bank of preset material properties using sparse, text labels in natural language. Our physics-based synthesis pipeline creates realistic movements that reflects either the underlying material, or during editing, the intent of the user.

\textbf{Realistic, Physic-Based Dynamics.} Using features from large-scale 2D vision models~\cite{kirillov2023-segmentanything,oquab2023-dinov2,radford2021-CLIP}, feature splatting is capable of not only segmenting objects for editing, but also ground physical properties of components within objects using language. In the second sequence, given the text prompt `a vase with flowers', the flowers sway as if it were blown with winds, but the vase remains still as it is considered as rigid materials by feature splatting. The fourth sequence emphasizes volume preservation, where the elastic ball is internally filled and bounces off the ground.

\textbf{Spatial consistency.} Conceptually, our method synthesizes views that are consistent across different camera viewports and timesteps of the physical simulations. For all sequences in Fig.~\ref{fig:physics_simulation_main}, feature splatting is able to synthesize 3D-consistent views. When objects are removed or deformed from their original places, the regions that were originally occluded are nicely synthesized.

\begin{figure}[t]
    \centering
    \includegraphics[width=\linewidth]{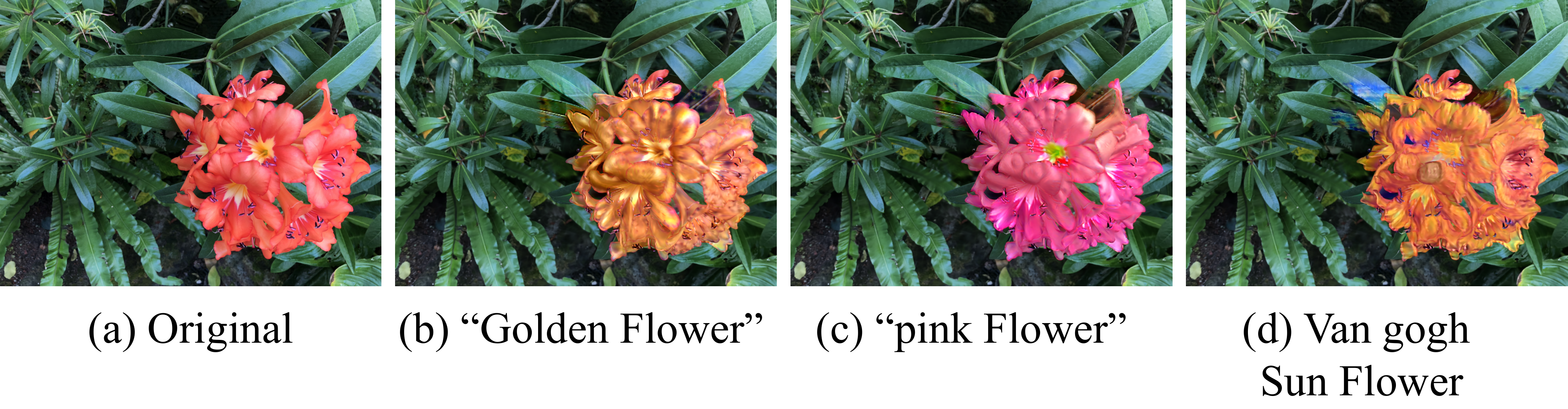}
    \caption{Language guided appearance editing. we optimize the SH coefficients via cosine similarity between CLIP embedded language and image feature.}
    \vspace{-1em}
    \label{fig:appearance_editing}
    % \vspace{-2em}
\end{figure}

\textbf{Temporal continuity.} Our physical editing approach leverages differentiable physics simulation engine~\cite{hu2019-taichi} and our volume preservation technique fills the interior of gaussians using transparent particles. Combined, our technique enforces a smooth and continuous transition from static scenes to object dynamics, where the rendered views are consistent with vanilla GS~\cite{kerbl2023-gaussiansplatting} at $T=0$. Compared to some previous scene-editing methods that use black-box generation models~\cite{haque2023-instructnerf}, our synthesis process offers both temporal consistency and great explainability.

\textbf{Real-time Efficiency.} Feature splatting maintains the ability to perform real-time rasterization, similar to Gaussian splatting~\cite{kerbl2023-gaussiansplatting}. The bottleneck of our physics simulation pipeline is the Taichi physics simulation engine, which runs at an approximate average of \emph{30 fps} on a desktop-grade GPU. Our feature splatting pipeline is optional during inference. Thus, with computed particle trajectory, images can be synthesized at approximately \emph{100 fps}.

\subsection{Editing Appearance and Geometry}

With the unified representation of feature splatting that hosts geometry, texture, and semantics, we demonstrate how to interact with objects in the scene via basic editing primitives and language-guided appearance editing.

\textbf{Geometric Editing.} 
We showcase geometry editing using feature splatting, highlighting object removal, scaling, rotation, translation, and cloning, as illustrated in Fig.~\ref{fig:editing_primitives_viz}. With our open-vocabulary scene decomposition design that fuses features from multiple pre-trained models, these geometric operations are fully automatic and the editing results are artifact-free, which is of great potential for practical applications.

\textbf{Appearance Editing.} 
Thanks to the unified representations in feature splatting, editing the appearance of object is also straightforward. In Fig~\ref{fig:appearance_editing}, we demonstrate the capability to edit the appearance of objects using CLIP guidance. We train a scene using feature splatting then update only the SHs of the selected Gaussian centroids using cosine loss between rendered image and CLIP with text prompt: ``A photo of an {\it Adj.} flower''. As demonstrated, feature splatting excels in both color and style editing while preserving the background in just 2,500 iterations.

\begin{figure}[t!]%
\begin{subfigure}[t]{0.25\linewidth}
\includegraphics[width=\textwidth]{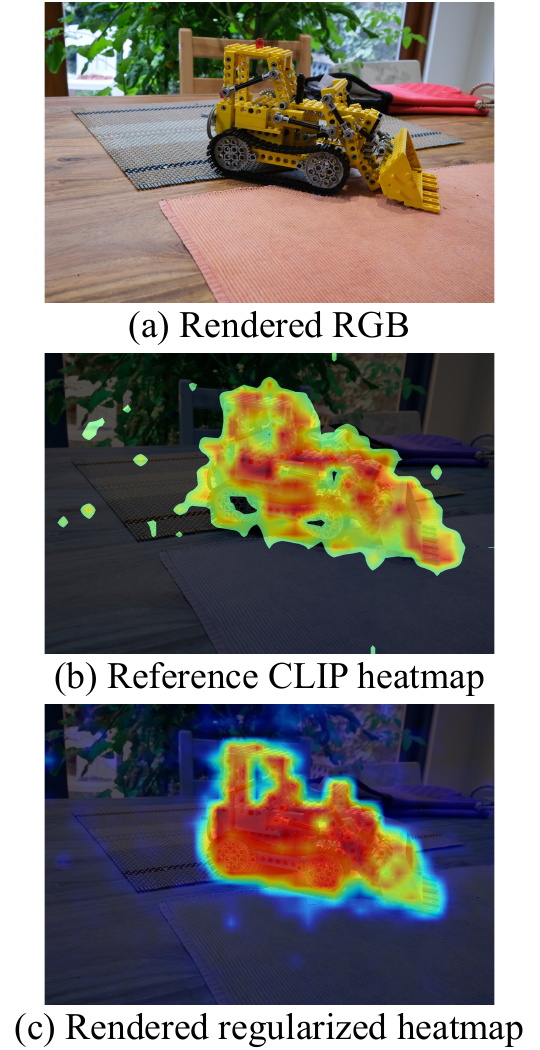}
\caption{Ablations of multi-model regularization. Without regularization, rendered features completely overfit to reference CLIP features with artifacts.}
\label{fig:rendering_regularization}
\end{subfigure}
\hfill%
\begin{subfigure}[t]{0.72\linewidth}
\includegraphics[width=\textwidth]{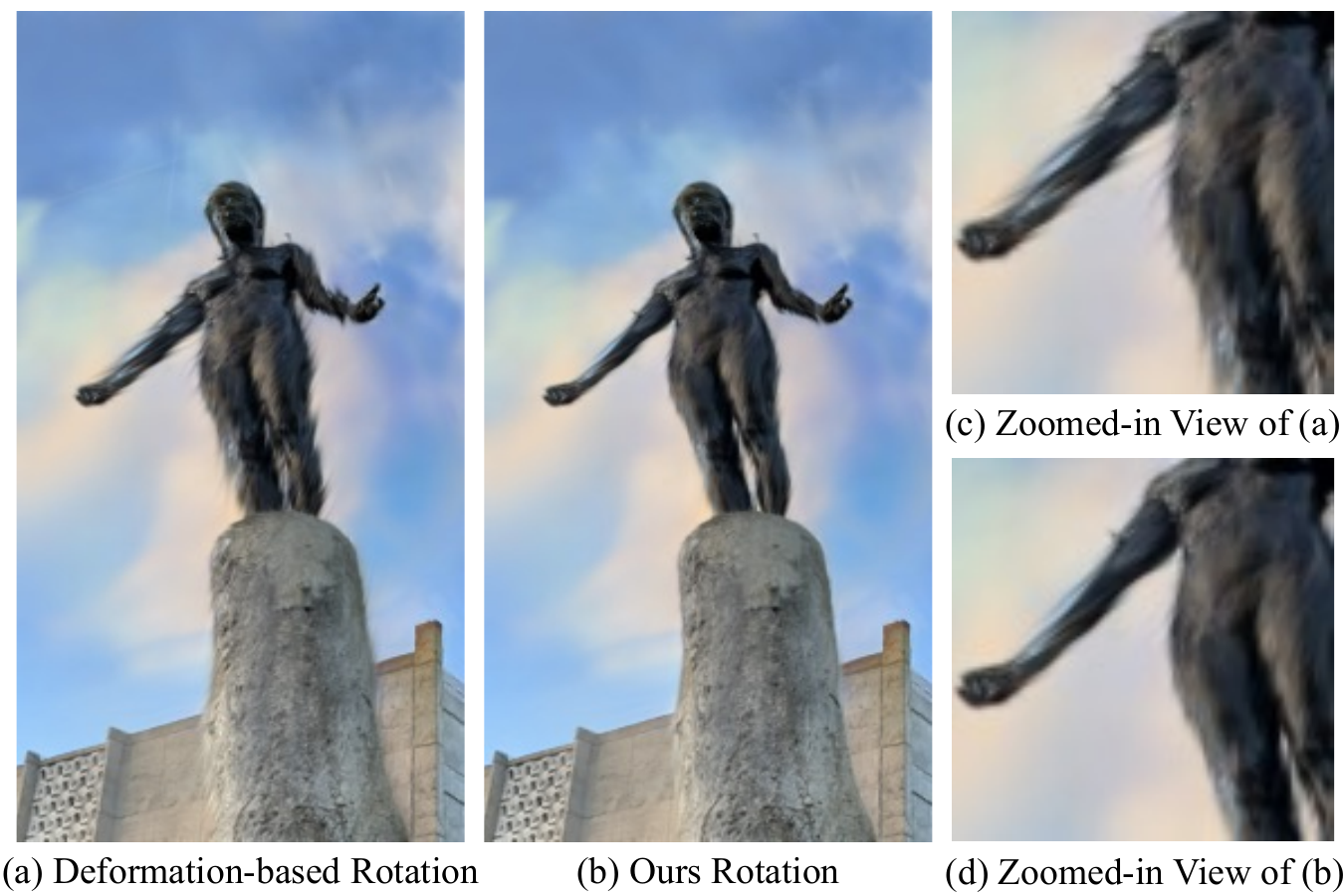}
\caption{Ablations of covariance rotation techniques. Though both methods perform similarly when the deformation is small, artifacts can emerge if we use rotations from deformation when the object undergoes large rotation and displacement.}
\label{fig:ours_rotation}
\end{subfigure}
\caption{Ablations of the regularization effects of fusing features from multiple pre-trained models and our rotation estimation technique.}
    \label{fig:combined_qual_ablation}
\vspace{-1em}
\end{figure}

\begin{table}[t]
\begin{minipage}{.57\textwidth}
  \caption{Timing ablation of our optimization techniques. Feature splatting reduces the training time by over 60\% compared to the baseline even when the feature dimension is the same. {\it Half2} stands for half2 compiler intrinsics, {\it GBuf} stands for shared gradient buffer, and {\it IMem} is interleaved memory access.}
  \label{table:cuda_optimization_ablation}
  \centering
 \begin{tabular}{c  c  c  c  l}
 \toprule
 Feat. Dim.  & Half2 & GBuf & IMem & Time
 \\
 \midrule
 \multirow{4}{*}{768} &{\color{red} - }&{\color{red} - }&{\color{red} - }& 3.21
 \\
 & {\color{green}\cmark} &{\color{red} - }&{\color{red} - }& 1.96{\scriptsize (-38.9\%)}
 \\
 & {\color{green}\cmark} & {\color{green}\cmark} &{\color{red} - }& 1.54{\scriptsize (-52.0\%)}
 \\
 & {\color{green}\cmark} & {\color{green}\cmark} & {\color{green}\cmark} & 1.21{\scriptsize (-62.3\%)}
 \\
 \midrule
 \multirow{2}{*}{256} & \multicolumn{3}{c}{Feature-3dgs~\cite{zhou2023-feature3dgs}} & 0.61
 \\
 & {\color{green}\cmark} & {\color{green}\cmark} & {\color{green}\cmark} & 0.41
 \\
 \midrule
 {\bf 32 (Ours)} & {\color{green}\cmark} & {\color{green}\cmark} & {\color{green}\cmark} & {\bf 0.06{\scriptsize (-97.1\%)}}
 \\
 0 & \multicolumn{3}{c}{Color-only GS~\cite{kerbl2023-gaussiansplatting}} & 0.05
 \\
 \bottomrule
\end{tabular}
\end{minipage}
\hspace{8px}
\begin{minipage}{.4\textwidth}
  \caption{Comparison of localization accuracy between OWL-ViT~\cite{minderer2022-owlvit}, LERF~\cite{kerr2023-lerf}, and feature splatting. Our method performs best on 2D localization and can directly localizing objects in 3D, which is not possible for LERF~\cite{kerr2023-lerf}.}
  \label{tab:lerf_localization}
  \centering
  \resizebox{\linewidth}{!}{
  \begin{tabular}{c  l  c}
     \toprule
     Loc. Space  & Method & Accuracy
     \\
     \midrule
     \multirow{3}{*}{2D} & OWL-ViT~\cite{minderer2022-owlvit} & 54.8\%
     \\
     & LERF~\cite{kerr2023-lerf} & 80.3\%
     \\
     & Ours-CLIP-only & 73.0\%
     \\
     & Ours-CLIP-DINO & 71.4\%
     \\
     & Ours (full) & {\bf 81.7\%}
     \\
     \midrule
     \multirow{2}{*}{3D} & LERF~\cite{kerr2023-lerf} & -$^*$
     \\
     & Ours & 50.7\%
     \\
     \bottomrule
    \end{tabular}
  }
\end{minipage}
\end{table}

\subsection{Ablations}
\label{sec:ablations}
We provide quantitative results on the impact of the system improvements, and ablation studies on our proposed techniques on feature splatting and the physics-based dynamic synthesis.

\textbf{System Optimization.} Here we ablate the efficiency of our engineering contributions. For baseline, we trivially extend the vanilla GS~\cite{kerbl2023-gaussiansplatting} implementation to render $N$-dim features instead of 3-dim RGB values. The results are presented in Table.~\ref{table:cuda_optimization_ablation}. Due to the prohibitively expensive memory requirement of baseline after densification, we compute the timing in the table by training only 1,000 iterations on provided sequences in the DB dataset~\cite{hedman2018-dbdataset}.

Our optimized implementation has better timing than our baseline implementation and Feature-3DGS~\cite{zhou2023-feature3dgs}. In comparison, feature splatting generally requires less than 1 hour on average for training, whereas Feature3DGS~\cite{zhou2023-feature3dgs} empirically measured to require 6 hours. We believe that the efficiency of our feature training implementation can facilitate future research.

\textbf{Fusing Features from Multiple Models.} In Fig.~\ref{fig:rendering_regularization}, we provide qualitative comparisons of heatmap generated by comparing text queries and rendered features to ablate the effectiveness of fusing multiple pre-trained vision models. We can see that the powerful ability of Gaussian splatting causes the guassian scene representation to {\it overfit} reference features, which results in the model fitting high-frequency details of reference features including imperfections in the coarse reference feature map. In comparison, with DINO regularization technique, the model learns a smooth feature representation that adheres to the boundary of objects, which is desirable. The overfitting without regularization is further quantitatively validated in Table.~\ref{tab:rendering_tab}, where the unregularized model has better CLIP feature rendering on training views, but worse on validation views.

\textbf{Rotating Gaussian Primitives.} We compare against the scheme proposed by~\cite{xie2023-physgaussian} in Fig.~\ref{fig:ours_rotation} without the additional regularization applied at modeling time. We observe that when the sculpture undergoes large deformation, the Gaussians rotated according to the deformation gradients technique from~\cite{xie2023-physgaussian} exhibit obvious surface artifact. Our approach produce fewer artifacts without additional processing, nor modification to the modeling procedure. %which validates the efficacy of our rotation estimation method.

\begin{table}[t]
\begin{minipage}{.4\textwidth}
    \centering
    \includegraphics[width=\linewidth]{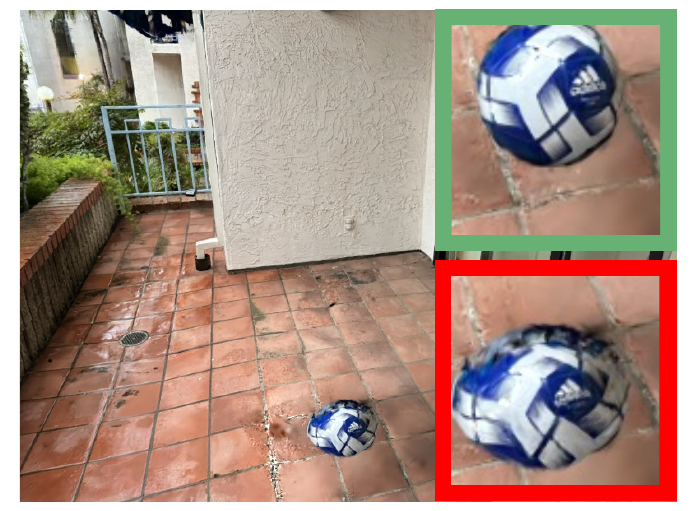}
    \captionof{figure}{Ablations of volume preservation. Without particles for internal support, elastic objects collapse and break upon collision (in red box); whereas correct simulated physics leads to a bouncing ball (in green box).}
    \label{fig:infilling_ablation}
\end{minipage}
\hspace{8px}
\begin{minipage}{.57\textwidth}
  \caption{Comparison of rendering quality and accuracy of rendered feature maps of feature splatting on collected datasets with 10\% holdout views. Conceptually, feature splatting does not interfere with color rendering. $\mathcal{L}_{\text{CTRAIN}}$ indicates CLIP feature loss for training views, $\mathcal{L}_{\text{CVAL}}$ for validation views.}
  \label{tab:rendering_tab}
  \centering
  \resizebox{\linewidth}{!}{
  \begin{tabular}{l c c c c}
     \toprule
     Method & PSNR$\uparrow$ & $\mathcal{L}_{\text{CTRAIN}}$ & $\mathcal{L}_{\text{CVAL}}\downarrow$ & FPS
     \\
     \midrule
     Vanilla GS~\cite{kerbl2023-gaussiansplatting} & 26.45 & N/A & N/A & 129
     \\
     Ours w/o DINO & 26.48 & {\bf 0.029} & 0.048 & 102
     \\
     Ours & 26.47 & 0.032 & {\bf 0.046} & 102
     \\
     \bottomrule
    \end{tabular}
  }
\end{minipage}
\vspace{-1.5em}
\end{table}

\textbf{Volume Preservation.} We ablate the effect of our implicit volume preservation method in Fig.~\ref{fig:infilling_ablation}. For the bouncing ball scene, if the ball is simulated only using the centroids of Gaussians without densified surface and internal support particle, the ball collapses upon collision with the floor under the effect of gravity, as shown in the image with zoomed-in view in the red box. On the other hand, our infilling technique enables correct physics where the ball bounces off the ground. (Better shown in videos on the project website).

\textbf{Localization.} In Table.~\ref{tab:lerf_localization}, we compare the 2D localization ability of feature splatting given text prompts. We use the localization dataset provided by LERF~\cite{kerr2023-lerf} and follow the evaluation protocol in LERF~\cite{kerr2023-lerf}. On the 2D setting, our full method achieves better accuracy than LERF, whereas removing DINOv2~\cite{oquab2023-dinov2} or SAM~\cite{kirillov2023-segmentanything} reduces the accuracy, validating the need to fuse features. This is consistent with our findings in Table.~\ref{tab:rendering_tab}. The performance drop when moving from 2D to 3D is expected because the scene includes many more candidates occluded from a single view. This introduces false positives not covered by the labels.

\section{Conclusion}

{\bf Limitation.} One limitation of feature splatting is artifacts in the background that may arise after object removal or displacement. Though this is a trivial extension from existing works~\cite{chen2023-gaussianeditor}, feature splatting currently does not perform inpainting after object removal, which may result in artifacts under certain circumstances.

In conclusion, we proposed feature splatting that enables language-guided interaction for Gaussian splatting. We demonstrate the success and effectiveness of feature splatting for physic simulations, geometry, and appearance editing.

% ---- Bibliography ----
%
% BibTeX users should specify bibliography style 'splncs04'.
% References will then be sorted and formatted in the correct style.
%
\bibliographystyle{splncs04}
\bibliography{main}

\clearpage

\appendix

\section{Language-Driven Appearance Editing via CLIP}

We apply classifier guidance and directly optimize the appearance of select objects using gradients from the vision model in CLIP. We start with a batch of rendered 2D images. Then to localize parts of the scene that we would like to change, we build a mask \(m\) by computing pixel-wise scores. We compute the gradient for each pixel via the cosine similarity loss between the embedding of the current view \(I\) and the text query \(L\):%
\begin{align}
    \mathcal L_\text{CLIP}(I, L)=1-\langle \mathrm{emb}_\text{img}(I), \mathrm{emb}_\text{text}(t)\rangle.
\end{align}
A notable benefit that Feature Splatting has over NeRF-based edits, is that local changes won't affect the entire scene. Distilled Feature Fields, for instance, require joint supervision with the original RGB images during editing, because modification of the neural network tends to affect the scene in a global manner. Therefore, conceptually, feature splatting converges faster for local edits since it updates only local parts of the scene.

\section{Implementation Details}

\subsection{Systems Considerations}

Naive attempts at storing additional high-dimension semantic features in 3D Gaussians lead to a dramatic slowdown in the modeling stage~\cite{luiten2023-dynamicgaussians}. Upon close inspection, we discovered that a sub-optimal memory access pattern is the bottleneck. The original Gaussian splatting code base writes the gradients directly to the global GPU DRAM during the backward pass. This parasitic effect is especially pronounced when the feature dimension is large because multiple gradient paths compete for limited access to the same DRAM slot. 

\paragraph{Gradient Buffer in the L1 Cache.} We resolve this problem by introducing a gradient buffer in the GPU L1 cache. We divide the images into 16 x 16 tiles. Each buffer has a size of \(256\), which means that each pixel can access this cache in an interleaved fashion. This significantly reduces the number of global memory access to the DRAM.

\paragraph{FP16 Tensors and Half\(2\) Arithmetics.} We replace the standard FP32 tensors with half-precision FP16 tensors, reducing the feature memory usage by half. We further improve by using the Half\(2\) intrinsic functions. This CUDA compiler feature `squeezes' two FP16 numbers into a single FP32 CUDA register, which further improves memory and arithmetic utilization.

We provide ablations on each of these techniques in the experiment. We plan to release our code as a reference implementation to facilitate future research that requires rasterizing any additional properties with Gaussian splatting.

\subsection{Staged Feature Splatting Pipeline}

The original Gaussian splatting trains for 30,000 iterations to capture all fine-grained texture details of the scene. However, for scene editing purposes, it is often redundant to optimize for such a long period for features. We empirically find that the semantics of large objects quickly emerge after a few hundred training iterations, and part-level semantics are well captured after a few thousand iterations. Therefore, to both improve the efficiency of feature splatting and to regularize features, we optimize features only for $N_{feat}$ ($N_{feat} = 2,500$). For new Gaussians that are cloned/split from existing ones during adaptive densification, we copy the features of the source Gaussians to the new Gaussians.

\subsection{Post-processing for Scene Decomposition}

Thanks to the explicit representation, we present two optional post-processing steps to reduce false positives and false negatives for this selection process, which is comparable to the {\it closing} and {\it opening} morphological operations. To avoid false negatives on object boundary, we apply a KNN operation to select non-selected Gaussians whose neighbors are mostly foreground objects. To filter out false positive noises, which are common on the boundary of the scene (such as the sky) where Gaussians are not well-reconstructed, we apply a clustering step using DBSCAN to filter out scattered Gaussians.

In addition, optionally, the user can include additional object-specific negative vocabularies for better differentiation. If such vocabularies are provided, we simply carry out the open-vocabulary segmentation described in the main paper for the additional negative vocabularies, and perform a set subtract to remove Gaussians selected by the negative vocabularies from the selected foreground Gaussians.

\end{document}